\documentclass[letterpaper]{article}
\usepackage{iccc}

\usepackage[utf8]{inputenc}

\usepackage{times}
\usepackage{helvet}
\usepackage{courier}
\usepackage{graphicx}
\usepackage{capt-of}
\newcommand{\outNim}[1]{}

\title{3D Topology Transformation with Generative Adversarial Networks}
\author{Luca Stornaiuolo$^1$, Nima Dehmamy$^2$, Albert-László Barabási$^2$,
\and Mauro Martino$^3$\\
$^1$Politecnico di Milano,
$^2$Northeastern University, 
\and
$^3$IBM AI Research\\
{\tt\small luca.stornaiuolo@polimi.it, \{n.dehmamy, a.barabasi\}@northeastern.edu, mmartino@us.ibm.com}
}

\begin{document}

\twocolumn[{%
\renewcommand\twocolumn[1][]{#1}%
\maketitle
\begin{center}
    \centering
    \vspace*{-1.1cm}
    \includegraphics[width=0.8\textwidth]{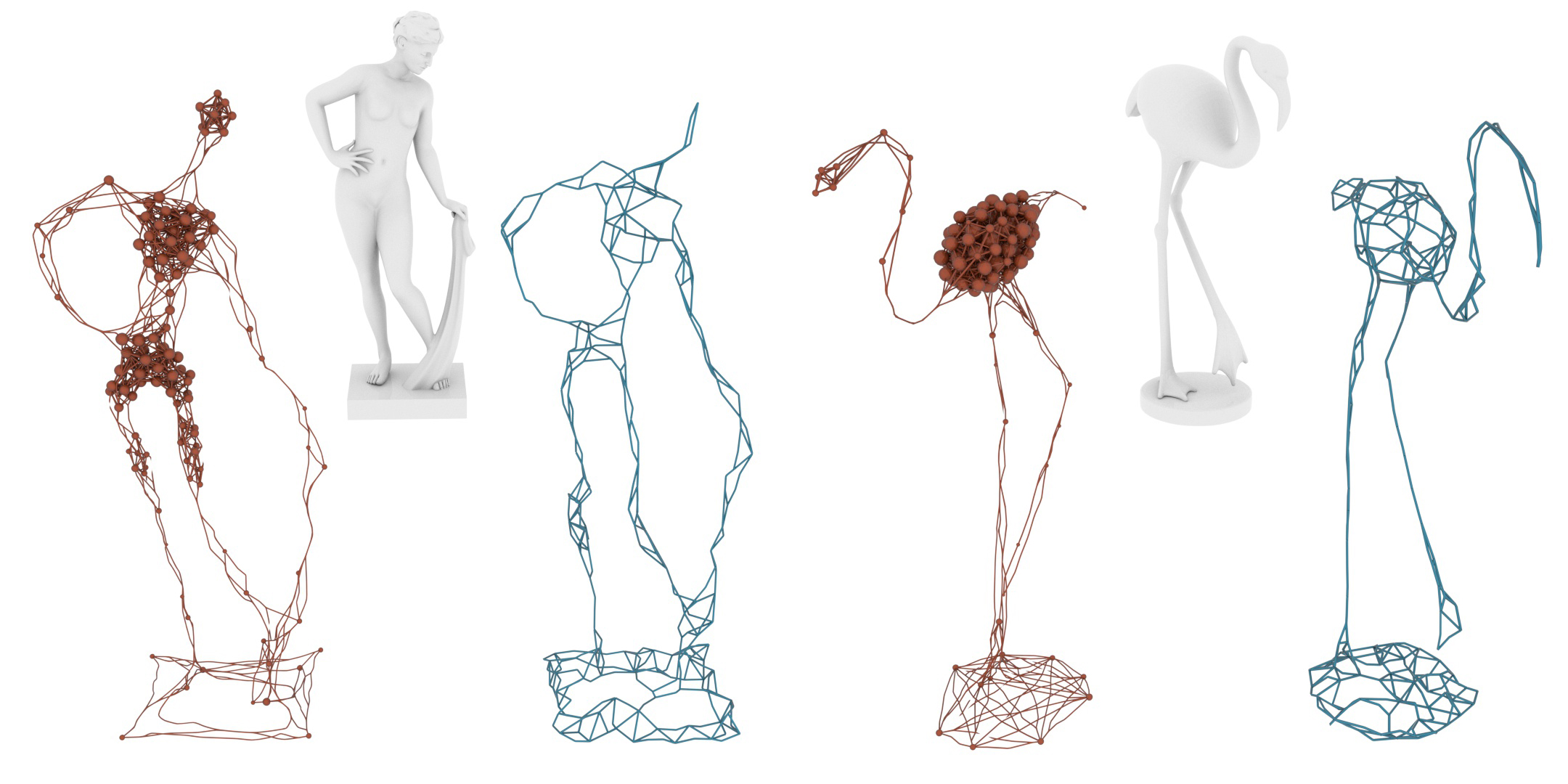}
    \captionof{figure}{
    \small
    Our proposed 3D Topology Transformation method disentangles the concepts of structural shape and volumetric topology style of objects in 3D space. 
    The Generator of our Vox2Vox is able to transform the 3D input models (on the top) to novel 3D representations (on the bottom), while retaining the original overall structure.
    }
    \label{fig:cover}
    \vspace*{0.8cm}
\end{center}%
}]

\outNim{

\maketitle

\begin{figure}[t]
    \centering
    \includegraphics[width =0.7\columnwidth]{figs/COVER_PAPER_vertical.jpg}
    \caption{
\small
    Our proposed 3D Topology Transformation method disentangles the concepts of structural shape and volumetric topology style of objects in 3D space. 
    The Generator of our Vox2Vox is able to transform the 3D input models (on the top) to novel 3D representations (on the bottom), while retaining the original overall structure.
    }
    \label{fig:cover}
\end{figure}
}

\begin{abstract}
\begin{quote}
Generation and transformation of images and videos using artificial intelligence have flourished over the past few years. 
    Yet, there are only a few works aiming to produce creative 3D shapes, such as sculptures. 
    Here we show a novel 3D-to-3D topology transformation method using Generative Adversarial Networks (GAN). 
    We use a modified pix2pix GAN, which we call Vox2Vox, to transform the volumetric style of a 3D object while retaining the original object shape.
    In particular, we show how to transform 3D models into two new volumetric topologies - the \textit{3D Network} and the \textit{Ghirigoro}. 
    We describe how to use our approach to construct customized 3D representations. 
    We believe that the generated 3D shapes are novel and inspirational.
    Finally, we compare the results between our approach and a baseline algorithm that directly convert the 3D shapes, without using our GAN.
\end{quote}
\end{abstract}

\section{Introduction}
Artificial Intelligence (AI) is experiencing enormous growth in popularity within the fields of creative arts and design. 
Image style-transfer based on deep neural networks \cite{gatys2016image,CycleGAN2017} 
has become very popular both in scientific and artistic communities, 
influencing image creation as well as videos and movies production \cite{huang2017real}.  
This approach allows achieving a decoupling of content and style in art \cite{johnson2016perceptual}, where AI has become able to generate new pieces of art using the style of one piece and the content of another.

In the context of utilizing machine learning algorithms to produce new artworks, Generative Adversarial Networks (GAN) have played an integral role in the recent studies, owing to their ability to learn representations of data and to generate outputs that mimic realistic elements \cite{radford2015unsupervised}, 
including songs, paintings, and  sketches
\cite{briot2017deep,dumoulin2017learned,ha2017neural}. 
While in many GANs the input is a random noise, 
conditional GANs (cGANs) \cite{mirza2014conditional} can be used to pair a specific input with the desired generated output, 
as in the case of the 
pix2pix architecture \cite{isola2017image}, able to generate new realistic images from different representations.
While most of these efforts in using GANs have focused on 2D images, the technology has become mature enough to apply these methods to 3D shapes.
Multiple works have generated realistic 3D reconstruction of shapes starting from photos \cite{wu2016learning,brock2016generative} 
Recently, there have also been many works on generating or learning a point cloud representation of 3D objects \cite{achlioptas2017learning,li2018point}. 
There have also been many works on converting 2D images or depth-maps to 3D \cite{li2019synthesizing,shin2018pixels}. 
However, there has been little research on how to create stylistically different 3D representations of a given 3D shape. 

In this paper, we present Vox2Vox, a new 3D cGAN, that generalizes pix2pix to 3D. As shown in Figure~\ref{fig:cover}, Vox2Vox is capable of transforming the topology and the internal structure of a 3D object, while maintaining its overall shape. 
We propose two volumetric topologies, the \textit{Ghirigoro} and the \textit{3D Network}, and we describe how to train our Vox2Vox model to create new topology transfers. We then demonstrate the effectiveness of our approach by comparing the results obtained with our AI model against those obtained with a pure procedural algorithm.


\section{Related Work}
Before proceeding to describe the details of our approach, we will review other work on creating 3D shapes with AI. 

\subsubsection{3D Shapes Synthesis.} 
3D shape synthesis 
is used to model and reproduce shapes in 3D space. 
While the first applications of 3D geometry models mainly concerned video games and visual media, nowadays, reproducing 3D shapes of real world objects influences 
design and architecture, to self-driving cars, 
to scientific and medical visualizations.
Our work contributes to a shift from the manual creation of 3D models with computer-aided design tools 
to automatically generated 3D shapes from different inputs. 
\outNim{
One possible approach is to \textit{use sensors} to obtain data on real objects and to reproduce them within the 3D space, combining different information. Examples of this approach are \cite{longuet1981computer, szeliski1994recovering, cui20103d, lanman2009build, furukawa2010accurate}, where it is possible to recover 3D shape from image streams of a camera, structured light projection, or laser stripe projection, by processing the information of the scanned real object. However, these techniques are limited to having a real object with the same shape of the desired output.
}
In addition to approaches that use probabilistic graphical models \cite{kalogerakis2012probabilistic}, the advent of Deep Learning brings the opportunity to automatically synthesize novel 3D shapes by \textit{assembling parts} of 3D objects, extracted from model databases, to create new compositions \cite{huang2015analysis}. 
However, 
these approaches require collections of \textit{labeled} 3D object parts, which despite the release of many 3D databases  are still hard to acquire. 
To relax the labels requirement, view-based generative models are employed. 
They allow for reconstruction of 3D shapes given 2D images that represent one or more \textit{view points} of the object \cite{su2015multi,qi2016volumetric,soltani2017synthesizing}. 
However, reconstructing 3D shape starting from its 2D representation often yields low quality results due to the missing information, making it hard to produce unique inputs and realistic 3D objects. 

\subsubsection{Generative Adversarial Networks.}
GANs have become one of the most popular neural network architectures to generate novel realistic outputs \cite{radford2015unsupervised,brock2018large}. 
GANs are composed of two neural networks, the Discriminator, trained to distinguish between the real and the generated inputs, and the Generator, trained to produce new outputs that look real. 
GANs have been used to generate \textit{voxel-based} object representations in the 3D space \cite{brock2016generative,wu2016learning}, starting from random noises. 
However, producing a desired shape requires exploring the latent space in order to find the correct random input for the Generator. 
In this context, cGANs introduce non-random inputs to control the outputs of the Generator \cite{goodfellow2014generative}. 
cGANs have been employed as 3D shape Generators in \cite{wu2016learning,xie2018learning}, where 2D images and a conditional probability density function  are respectively used to condition the output of the Generator and 
 map the input to the desired 3D shapes. 
However, only few works exist about controlling the output of the Generator with 3D shapes as conditional inputs. 
\cite{ongun2018paired} presents a 3D-cGAN able to perform rotations of volumes in the 3D space. 
In comparison, our approach aims to transfer the shape of the input, while modifying its volumetric topology to obtain a novel collection of 3D objects, and, to the best of our knowledge, our is the first 3D-cGAN model for 3D topology transformations.

\subsubsection{Style Transfer.} Style transfer aims to learn the content and the structure of an input element while changing its style based on a different style source. 
In the past, many works have been released that perform 2D-to-2D style transfer on images and videos \cite{johnson2016perceptual,gatys2016image}.
2D-to-3D style transfer is mainly employed to apply specific textures or colors from 2D samples to 3D objects \cite{nguyen20123d}. \cite{kato2018neural} presents a novel 2D-to-3D style transfer approach able to perform gradient-based 3D mesh editing operations to modify also the surface of the 3D shapes based on the image used as style source. 
\cite{ma2014analogy} is one of the first works about 3D-to-3D style transfer: the proposed algorithm computes the analogy between one source element and the related target and applies it to synthesize new outputs based on different sources. 
This deformation-based approach is also used to generate different poses of animal meshes \cite{sumner2004deformation} and modify the design of 3D furniture, buildings models, and different classes of objects \cite{xu2010style,liu2015style,lun2015elements,mazeika2018towards}. 
However, these works require a specific formulation of analogy between the different parts of the analyzed objects that limits their application to few collections of 3D models.
In this context, our work lays the foundation for a novel 3D-to-3D style transfer. Indeed, with our 3D-cGAN model for 3D topology transformations, we can say that the style is sedimented in the trained network and multiple styles are supported with different weights of our Vox2Vox Generator. A future step would be allowing both shapes and styles as inputs of the Generator.
\begin{figure*}[t]
\vspace{-10pt}
    \centering
    \includegraphics[width=\textwidth]{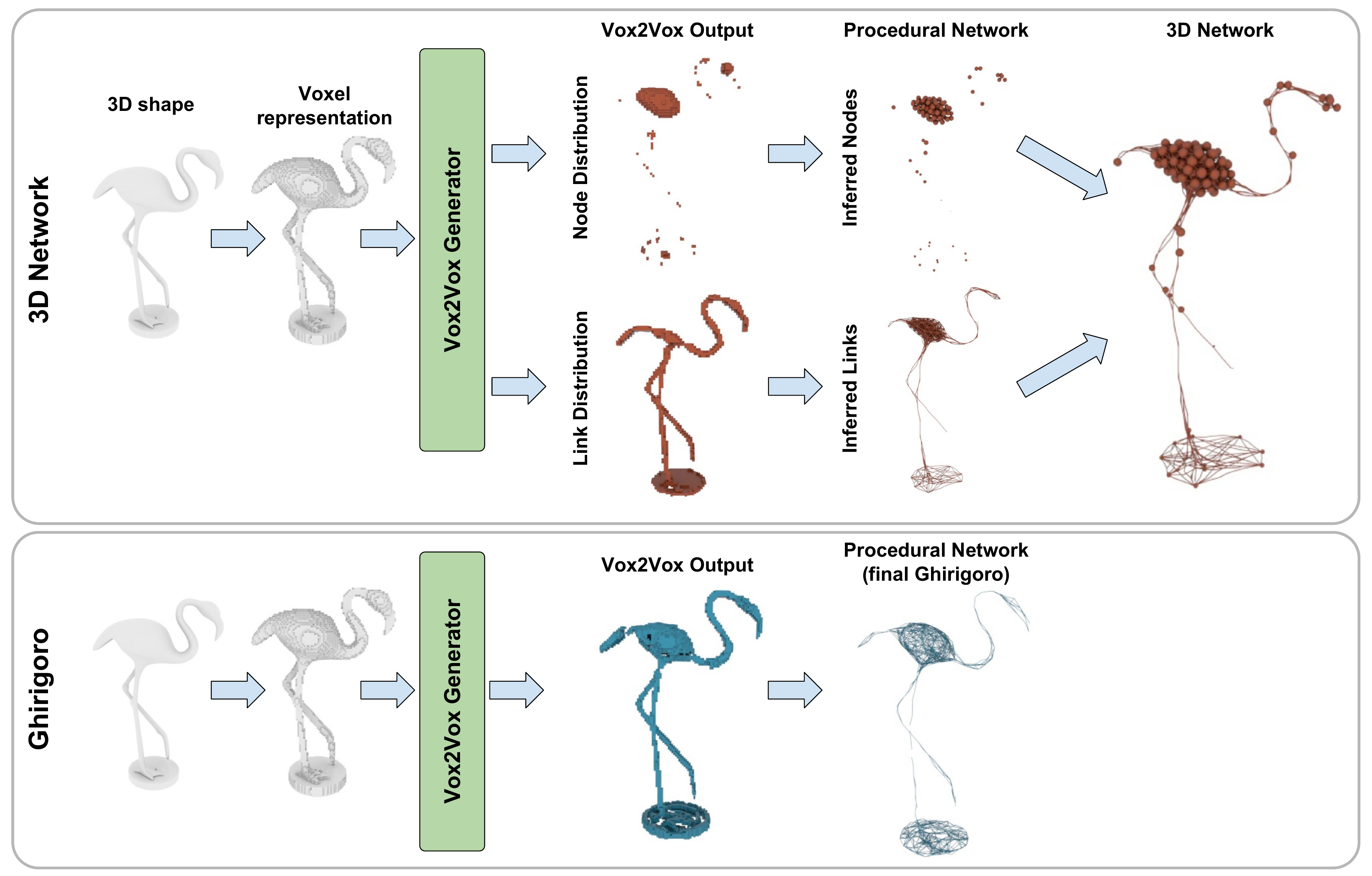}
    \caption{
    \small
    \textbf{Pipeline.} We present two 3D representations: the \textit{3D Network} topology (top) and the \textit{Ghirigoro} topology (bottom). 
    Our pipeline starts with a \textit{3D model} and converts it to its filled \textit{3D voxel representation}. 
    The 3D voxel representation is fed as input to the Generators of Vox2Vox trained to perform the topology transfers. 
    The results are the \textit{new 3D voxel representations}, where different channels contain different information of the new 3D shape (e.g. node and link distributions and for the 3D Network topology). 
    Finally, the \textit{output 3D model} is reconstructed from the voxel representation with our Procedural Network algorithms.
    }
    \label{fig:pipeline}
\end{figure*}

\section{Approach}
The goal of this paper is to convert a 3D shape into an alternative 3D representation inspired by the original. 
We will focus on converting the shape into a \textit{3D Network}, which is an abstract representation
in the same vein as cubism in art.
A network (graph) consists of two types of components: 1) Nodes, being the entities to be connected and 2) Links, which are the wires connecting the nodes. 
Like scaffold that supports the interior of the 3D object, we want the nodes to be placed in suitable locations inside the shape and links to connect these nodes to form a 3D Network. 

However, as we will discuss later, a random or a uniform choice of where to position nodes and link points doesn't result in aesthetically pleasing results.
Thus, we have to find a better way to choose where to place the nodes and link points. 
This requires us to find what part of the 3D shape should contain more nodes in order for the shape to be represented better than random nodes, and we would have to hand-code the criterion for finding this node density. 
If, however, the aesthetics is changed slightly, e.g. having curvy links like a scribble taking the shape of the network, we will need to, again, hand-code a procedure for the new aesthetics. 
Solving this inverse-problem of finding an algorithm or rule for suitable distribution of node and link positions inside the 3D shape can be very difficult. In contrast, training a system that would learn by examples could be much easier.
Therefore, we want our Vox2Vox cGAN to be able to produce the node and link distributions, using voxels. Once we have this distribution, we can use a procedural algorithm, we called \textit{Procedural Network}, that takes the voxels as an input (together with the number of nodes and the number of points to be used for links connecting nodes) and generates the final 3D Network representation.

In this way the main challenge moves from creating an algorithm that performs the transformation of the desired topology (in this case Network 3D) to creating the dataset necessary for training the AI model. 
Fortunately, creating arbitrary 3D Networks and converting them to space-filling 3D blobs is much easier than doing the reverse, as we will discuss in the section about model training. 
\outNim{
Note that the problem of creating a particular 3D Network is easy, when the large-scale structure is not constrained.
But the reverse problem of constructing a 3D Network constrained to a particular 3D shape is more involved. 
This can be compared with the following analogy in painting. 
It is easier to randomly splatter colors on a canvas, than it is to construct Pollock painting with splatters occurring in strategic points on the canvas to illustrate a particular shape. 
Yet, if one has experienced many versions of Pollock-style paintings one can combine the elements learned from them to convert a sketch into a Pollock-style painting. 
Additionally, in cases such as architectural design, features need to be distributed in a certain way inside the object to ensure stability, among other things. 
}
Once we have 3D Networks with their respective 3D figures, the 3D-cGAN is able to learn how to convert the latest to the former, while finding a pure procedural algorithm for doing so may be difficult and an optimization process may be costly. 
\outNim{
Based on this observation, we take the approach of creating a training dataset that contains many examples which start from a 3D Network and also convert it to a 3D shape. 
This way an AI can learn the correlation between the shape and the internal network structure, and apply this knowledge to converting other shapes into networks.
}
Another benefit of this approach is that, if we want to choose a different aesthetics, we can create a new aesthetics-related training dataset and not worry about the inverse problem.

\section{Proposed Method}
As stated in the previous section, solving the problem of creating a neural network to perform topology transfer is equivalent to creating a neural network that can take a 3D shape and produce a voxels-based distribution of nodes and links to put inside it. 
Thus, the output should preserve the general 3D structure of the input. 
It was shown in \cite{isola2017image} that when preserving the overall location of features in the input and output is important, the U-Net architecture performs much better than an encoder-decoder architecture, owing to its \textit{skip layers}. 
U-Net \cite{ronneberger2015u} is also known to converge with relatively little training data, making it ideal for our application. 
We, therefore, choose a U-Net architecture for the main component of our neural network. 
However, given a 3D shape, there are many different ways that one can fill this shape with a network. 
This means that when generating 3D blobs from networks, many different network layouts may end up having similar 3D blobs. 
This will confuse a neural network, as the same input is being assigned different 3D Networks as ``label''. 
This is precisely where a cGAN becomes useful. 
Unlike a simple neural network, where multiple labels would result in the network choosing the average of those labels, in a GAN the loss is minimized as long as the GAN learns to produce one of the correct labels, so that the Discriminator cannot tell the output from the real data. 
In summary, we need a cGAN which has U-Net architecture for its Generator and a Discriminator suitable for classifying 3D shapes. 
This is essentially exactly what the pix2pix  architecture \cite{isola2017image} does, only in 3D instead of 2D. 
Thus, the architecture of Vox2Vox will be very similar to pix2pix, but the 2D convolutional layer will be replaced with 3D convolutions and the number of layers and filters will be different.
\outNim{
Note that the output of the AI has spatial correlation with the original shape, but only in the high-level features.  
Thus, the cost function must take the original and the generated shape and compare them at high-level.
It must also compare the low-level features appearing in the generated  shape with those in the desired aesthetics to ensure style similarity. 
Since we want to preserve the large-scale 3D structure of the shape, most of the Generator's architecture must consist of 3D convolutional layers. 
To have the ability to compare the original and the new representation, we choose the pix2pix cGAN \cite{isola2017image} as the basis of our 3D-cGAN. 
We perform some post-processing to infer and extract a graph from the Generator output. 
}

\subsubsection{Pipeline Overview.}
Figure~\ref{fig:pipeline} shows an overview of the pipeline for converting a 3D shape into a 3D Network. 
First we transform a 3D input mesh to a filled voxel representation, given a target 3D resolution. 
The filled voxels are passed to the Vox2Vox Generator which outputs two channels of voxels: ch. 1 for the distribution of where to put nodes and ch. 2 for distribution of points along links connecting the nodes.  
\outNim{
The third component is composed by our 3D-cGAN. Based on the desired output topology style, a different pre-trained Generator is loaded and the filled voxel passes through the network. 
The output of the Generator is a new voxel representation of the input model and may use different channels to feature different information of the new topology (eg. two channels are used respectively to represent nodes and links for the 3D Network one). 
}
Finally, the voxel distributions for nodes and links are passed to the Procedural Network (described in detail below) algorithm to produce the final 3D Network. 

\subsubsection{Model Architecture.}
\outNim{
As claimed before, our 3D-cGAN is inspired by the pix2pix architecture extended to 3D.
We chose this architecture because we want the output to mimic the overall shape structure of the input. 
Since our pix2pix-inspired network deals with voxel representations, we use to call it \textit{Vox2Vox}.
The power of such architecture is that we can teach it to convert shapes into any abstract representation of them, as long as there is spatial correlation between the two 3D shapes.
}

\begin{table}[t]
\centering

\scalebox{0.9}{
\begin{tabular}{|lcccc|}
\hline
\textbf{Generator}  & \begin{tabular}{c} Training \\ shape
\end{tabular}
& \begin{tabular}{c} Predict \\ shape
\end{tabular} & Filters & Params \\ \hline
Input Layer  & $ 64^3 $ & $ 192^3 $ & $  1$ & - \\ 
Down3D*  & $ 32^3 $ & $ 96^3 $ & $  32$ & 2,080 \\ 
Down3D  & $ 16^3 $ & $ 48^3 $ & $  64$ & 131,392 \\ 
Down3D  & $ 8^3 $ & $ 24^3 $ & $  128$ & 524,928 \\ 
Down3D  & $ 4^3 $ & $ 12^3 $ & $  128$ &  1,049,216 \\ \hline
Up3D  & $ 8^3 $ & $ 24^3 $ & $  256$ &  1,049,216 \\ 
Up3D  & $ 16^3 $ & $ 48^3 $ & $  128$ &  1,048,896 \\ 
Up3D  & $ 32^3 $ & $ 96^3 $ & $  64$ &  262,304 \\ 
Up3D*  & $ 64^3 $ & $ 192^3 $ & $  C$ & 8,194 \\ 
\hline
\end{tabular}
}
\\ ~
\\
\scalebox{0.9}{
\begin{tabular}{|lccc|}
\hline
\textbf{Discriminator}  & \begin{tabular}{c} Output \\ shape \end{tabular} & Filters & Params \\ \hline
Input Layer     & $64^3$ & $C$ & - \\ 
Down3D*         & $32^3$ & $1$ & 3,088 \\ 
Down3D          & $16^3$ & $3$ & 32,800 \\ 
Down3D          & $8^3$ & $64$ & 131,136 \\
Down3D$^+$*         & $8^3$ & $1$ & 32,769 \\
\hline
\end{tabular}
}
\\ ~
\\
{\scriptsize * No Batch Normalization $^+$ Kernel size=8}
\\
\caption{
\small
\textbf{Model Architecture.} The Vox2Vox Generator has a U-Net architecture  
 made of \textit{Down3D} modules which contain 3D convolutional layers, 
and \textit{Up3D} modules which does an Upsampling by a factor of 2 in each direction, followed by a 3D convolutional layer, Dropout and Batch Normalization. 
The Discriminator consists only of \textit{Down3D} modules.
The number of filters, output dimensions and the number of parameters for each layer in the Generator and Discriminator are reported. 
}
\label{tab:gen}
\vspace{-15pt}
\end{table}

\begin{figure}[t]
    \centering
    \includegraphics[width =\columnwidth]{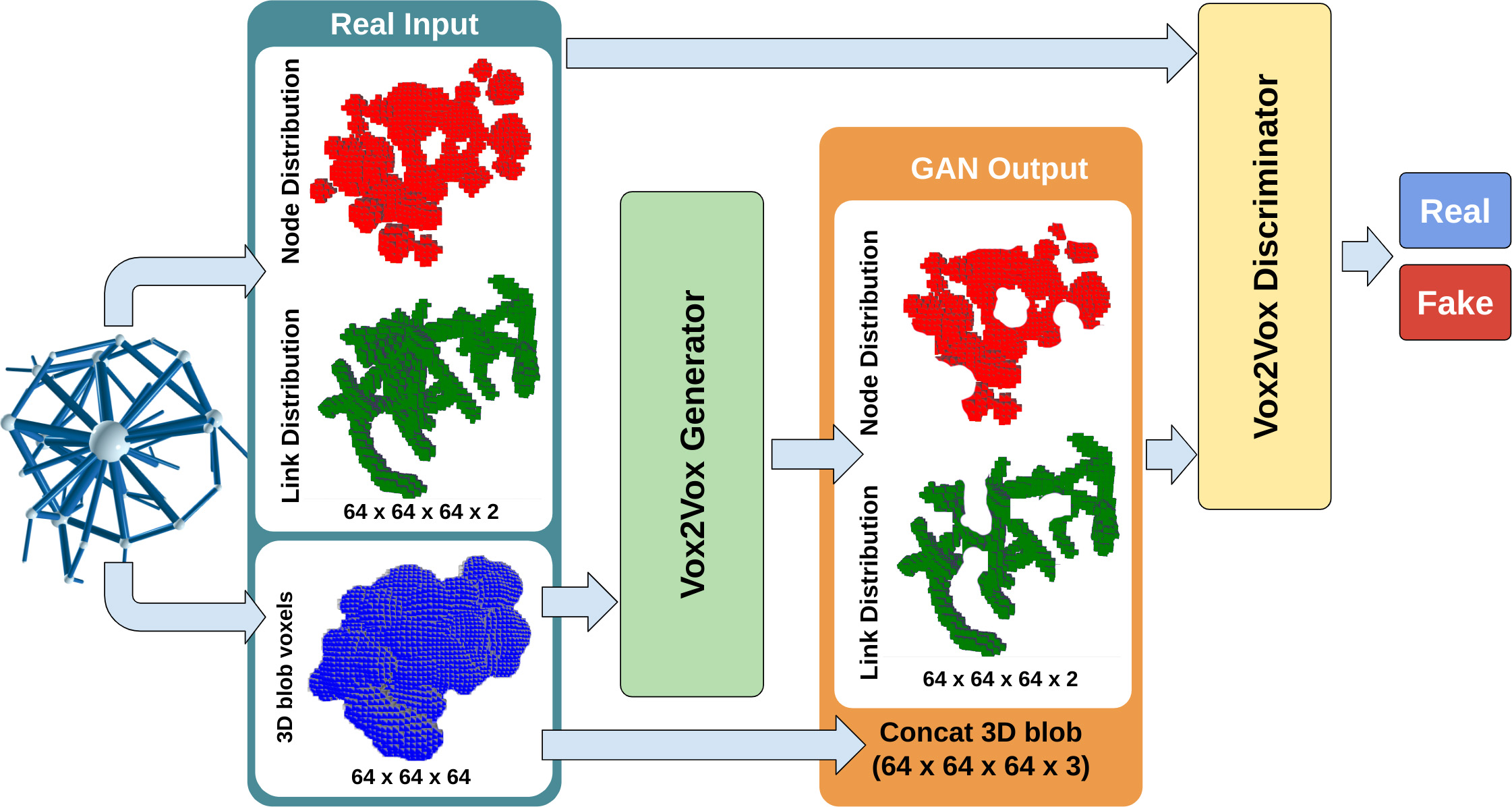}
    \caption{
\small
\textbf{Training Vox2Vox:}
    For conversion to a 3D Network, we first create a network and lay it out in 3D. 
    Then we convert the network into two formats: 1) a \textit{3D blob} found by assigning large radii to nodes and large thickness for links and merging them together (input of the Generator); 2) Separate \textit{node and link distribution} data with nodes having reasonably small radii and links being relatively thin (real input of the Discriminator). 
    The output of the Generator will have two channels, the predicted node and link distributions (fake input of the Discriminator). 
    }
    \label{fig:vox2vox}
    \vspace{-10pt}
\end{figure}

Table \ref{tab:gen} shows the details of the modules used in the Vox2Vox Generator and Discriminator. 
The Generator has a U-Net architecture, with four \textit{Down3D} modules in the encoder and four \textit{Up3D} modules in the decoder each of which contains a 3D convolutional layer. 
During training, the input shape is $64 \times64 \times64$ and the encoder layers have $32,64,128 $ and $128$ filters, encoding the input to $4\times 4\times 4\times 128$ shape. 
The decoder layers have $128,64,32$ and $C$ filters.
$C$ is the number of desired output channels, which for the 3D Network problem is two, one for node and one for link distribution. 
The Up3D modules also get the output of the encoding layers as input, as in U-Net. 
The Discriminator consists of four 3D convolutional layers, yielding a $8\times 8\times 8$ output. 
The cost function for the Discriminator decides whether on all the $8\times 8\times 8$ patches of the node and link distributions match the input 3D blob in a certain way. 
If the Generator successfully produces a node and link distribution which matches the blob over all the $8\times 8\times 8$ similar to how the real data matches the blob, the Discriminator will get fooled.
Because of the third dimension, the training process of Vox2Vox is both very memory intensive and computationally very expensive. 
That is why we chose to do the training in the resolution of $64 \times64 \times64$ voxels.
However, during prediction we increase the input size to produce higher quality outputs.
\begin{figure*}[t]
    \centering
    \includegraphics[width=0.98\textwidth]{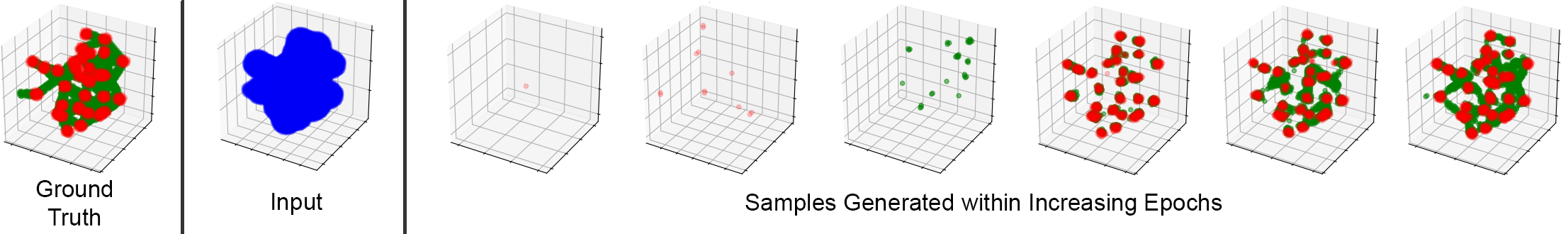}
    \caption{
\small
    \textbf{Model Training.} Sequence of 3D Network outputs from the Vox2Vox Generators while growing of epochs.
    }
    \label{fig:trainingdata}
\end{figure*}
Indeed, a nice feature of U-Net is that, since all layers are convolutional, we can produce larger outputs by simply modifying the input shape of the trained network. 
Note that, this increase in input size will not result in larger ``features''. 
For example, if the Vox2Vox trained at resolution $64$ produces nodes with radius $5$ voxels maximum, increasing the input shape to $192$ will still only produce nodes of radius $5$. 
Similarly, if at resolution $64$ the maximum length of links produced is $10$ voxels, it will be the same when the input size is changed to $192$. 
This is because the convolutional layers in all layers will still have the same receptive field of, say $8\times8\times8$ on the input, at both $64$ and $192$ resolutions. 
However, what this allowS us to do is that we can feed arbitrary large inputs to a trained Vox2Vox ($192$ was the maximum size our GPU memory allowed) and convert them into 3D Networks. 
Moreover, making the input larger allows us to convert objects with much higher details than the size of the objects in the training set. 
Therefore, after we train Vox2Vox on $64\times 64 \times 64 $ voxels, we change the input shape to $ 192\times 192 \times 192$
and produce higher quality outputs. 
We used Vox2Vox on a variety of different 3D sculptures and passed the output to the Procedural Network Algorithm to extract the generated 3D Network (Fig. \ref{fig:vox2vox}). 

\begin{figure*}[t]
    \centering
    \vspace{-5pt}
    \includegraphics[width=\textwidth]{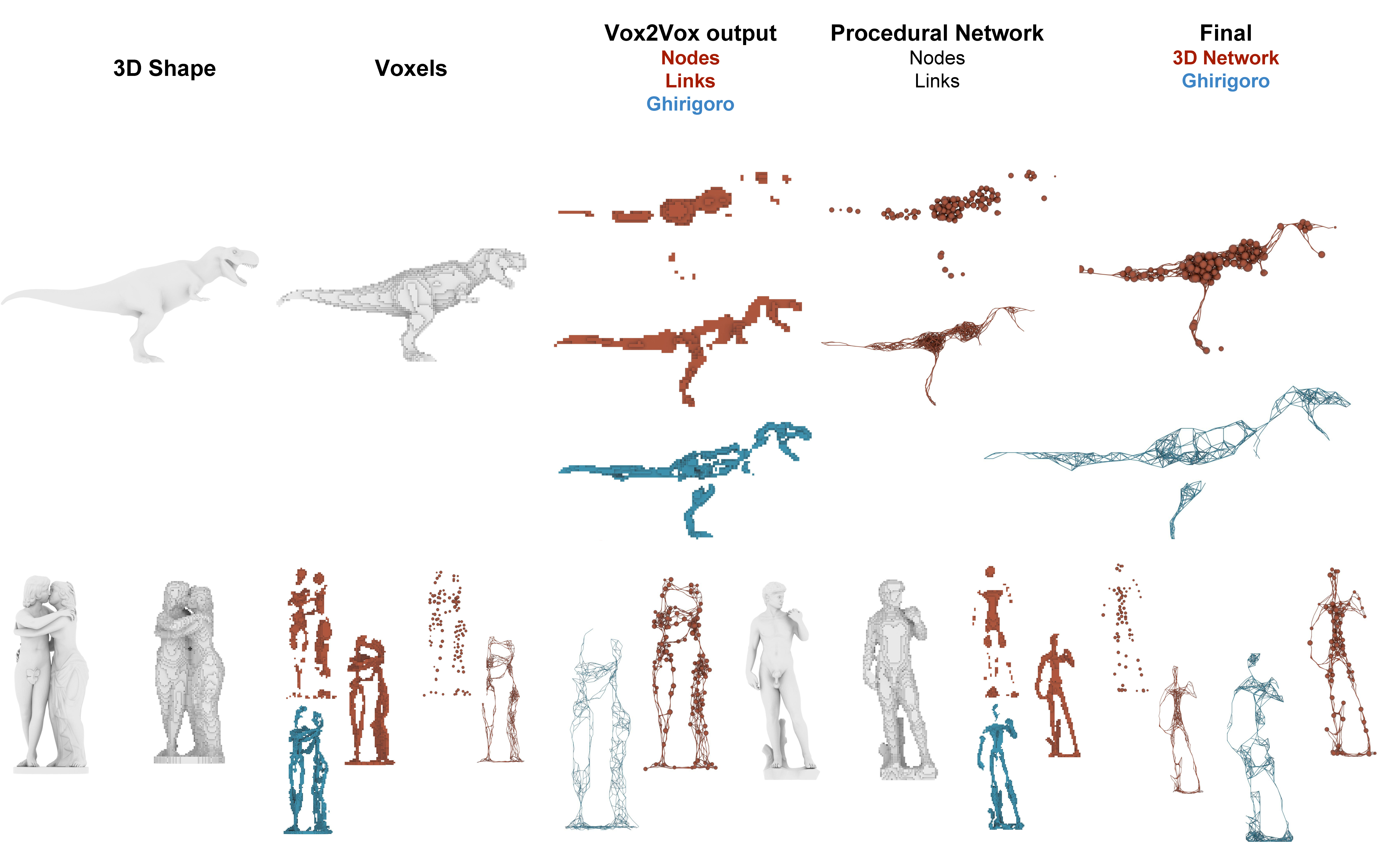}
    \vspace{-10pt}
    \caption{
    \small
    \textbf{Results.} Example of results for the 3D Network and the Ghirigoro topologies. 
    The columns are similar to the pipeline illustration \ref{fig:pipeline}. 
    Red models refer to the 3D Network topology, and blue models are for the Ghirigoro.
    The First column shows the original 3D model, second the voxel representation, third is the Vox2Vox output, fourth the inferred network components and fifth is the final outcome.
    }
    \label{fig:results-multi}
        \vspace{-10pt}
\end{figure*}

\section{Model Training} \label{sec:training}
The training procedure of Vox2Vox (Figure \ref{fig:vox2vox}) is similar to the pix2pix one: 
we present the Discriminator with a pair of inputs, which is the concatenation of the $64\times 64 \times 64 $ array of the 3D shape with the $64\times 64 \times 64 \times C$ array of real or fake output along the channels, yielding a $ 64\times 64 \times 64  \times C+1$ array. 
For the real pair, the label is a $4\times 4 \times 4 $ array of ones, and for the fake pair the array is zero. 
The main difference is our training data is 3D and  completely synthetic (see sections about training data below).
Some points to note is that, we exploited the convolutional nature of the U-Net layers also during training. 
During the training process, the input is a $64\times 64 \times 64 $ binary array (0 in empty spaces and 1 inside the 3D shape) and output of the Generator is a  $64\times 64 \times 64 \times C$ array, where $C$ is the number of channels. 
While this training converges to reasonably good results after around 10 epochs, we found that first training on $32\times 32 \times 32 $ binary arrays and very small networks, and then doing transfer learning by increasing the input resolution resulted in faster convergence to reasonably good results.  


\subsubsection{3D Network Training Data.}\label{sec:training-data}
We create a variety of networks first and then lay them out in 3D using a custom force-directed layout algorithm, inspired from \cite{dehmamy2018structural}, which makes sure nodes do not overlap. 
We then create two arrays from the network. 
One is a $64\times 64 \times 64 $ array of the 3D shape of the layout, merging nodes and link segments by replacing them with overlapping large and small spheres, respectively. 
This 3D shape is the input of the Generator. 
The second is a $64\times 64 \times 64 \times 2$ array, with the first channel being the nodes and the second the links, which are again replaced by spheres, this time with smaller radii (Figure \ref{fig:vox2vox}). 
Then, we replace nodes with spheres and segments along links with spheres which overlap and make the connected links. 
We generate about 700 such networks and heavily augment the dataset by rotating them in multiples of 20 degrees about the x,y and z axes to create a dataset of about 30,000 data points. 
We trained Vox2Vox with a few different thicknesses for links and sizes for nodes to determine a good choice for the sizes of the nodes and links so as to avoid space-filling links and nodes, which would let the Generator exploit this structure and fool the discriminator by simply filling the inside of the shape. Figure \ref{fig:trainingdata} shows a sequence of 3D Network outputs from the Vox2Vox Generators while growing of epochs.

\subsubsection{Note on Input Network Topology.}
There exist many different generative processes for producing networks which result in very different connectivity patterns. 
We found that using random graphs \cite{erdHos1960evolution} resulted in poor results. 
In these graphs, any node is equally likely to connect to any other node, resulting in a completely random network, with all nodes having a similar degree (i.e. number of links attached to them). 
The Vox2Vox Generator was able to produce reasonable positions for the nodes, but failed at producing good links between them. 
Our hypothesis is that, since links in these networks are completely random, the Discriminator overfitted to the few samples it had been trained on and was never satisfied with any other variation of the links. 
In contrast, when we trained the Vox2Vox on networks generated using a ``rich gets richer'' (Barabási-Albert (BA))  model \cite{barabasi1999emergence} the results improved dramatically. In the BA model some nodes, known as hubs, have significantly more links than other nodes. 
The BA model has ``hubs'' which are nodes with a much higher degree than most other nodes, which contrasts it strongly from the random ER network where all nodes have more or less the same degree. 
We chose the size of the nodes in the 3D array to be a function of their degree. 
When making the voxel representation, we assign lager node radii nodes with a higher degree.  
We believe that this predictable correlation between the node size and the density of links was learned by the Generator and was exploited to fool the Discriminator, allowing it to produce good results.

Usually, to train GANs a single input is presented to the network at a time \cite{goodfellow2014generative}, meaning batch size 1. 
This will force the Generator to try to learn the features of a single example. 
However, this will result in a very slow progression of the training. 
On the other hand, larger batch size yields faster convergence, but presenting a large batch of data result in more of an averaged output, stopping it from choosing a single pattern.  
Therefore, in the initial stages of the training, we set the batch size to 8 to allow for a faster approach towards good filters. 
In later stages, we reduced the batch size to 2 and then 1 to fine-tune the results.

\textbf{Ghirigoro: Training on a Second Topology.}
Aside from converting shapes to 3D Networks, we also tried a second, related topology. 
The ``Ghirigoro'' (doodle) topology consisted of converting a 3D shape into a long scribble that mimics its shape. 
This style has only one output channel, which is a set of long curvy lines, crossing more rarely than the 3D Network case, and there are no nodes. 
To construct the final doodle shape from the GAN we used the Procedural Network algorithm with different settings. 
The result is not exactly a doodle, as the Procedural Network may cross-link two parallel pieces of the doodle, resulting in a network. 
Nevertheless, the aesthetic is close to a doodle. 
The pipeline is shown at the bottom of Figure \ref{fig:pipeline} and the results are presented along with the 3D Network case in Figure \ref{fig:results}. 

\section{Procedural Network algorithm \label{sec:2net}}
We interpret the two output channels of the Vox2Vox Generator as node and link distributions. 
To extract a network from these distributions, we first use K-means to choose centers for nodes and for points along the links. 
We choose a higher K for link points than nodes in the K-means.   
We then form a network from points that are close to each other, connecting link points and node centers in a dense network. 
To avoid connecting points that are far from each other, we partition the space into small regions and connect points within each partition to each other. 
To avoid artifacts from the position of partition walls, we repartition the space multiple times by slightly shifting the partition walls by a random value. 
This method of connecting nodes and link points works better than a k-nearest neighbor method, as all nodes need not have the same degree, and hence neighbors. 
This results in a series of connected paths connecting nodes using via intermediate link points. Finally, we use the A* search algorithm \cite{hart1968formal} to extract links between nodes. 


To see how using the Vox2Vox output as a prior for node and link distribution compares against not using having this prior, we apply the above procedure to the original 3D shape voxels. 
This means that, to find node locations we just take the filled voxels of the 3D shape and apply K-means to it. 
This will fill the interior of the shape with randomly chosen node locations. 
Our default was choosing 300 node locations. 
We then do the same thing for link points, except that we choose 50 times more points to serve as points along links (i.e. $50\times300$ points). 
The rest of the procedure is the same. 
Comparing the outcomes, we observe that the baseline procedure results in curved and broken links, while the proposed Vox2Vox model produces 3D shapes with more effectively distributed links and nodes. 
\section{Results}
Figure~\ref{fig:results-multi} shows the results of the proposed 3D topology style transfer approach. 
For each 3D input model, we present its volumetric transformations into the \textit{3D Network} topology  and the \textit{Ghirigoro} topology. 
We then compare in Figure~\ref{fig:results} the results of the baseline pure procedural algorithm that applies the volumetric topology transformation directly on the 3D input.
\begin{figure}[t]
    \centering
    \includegraphics[width=0.5\textwidth]{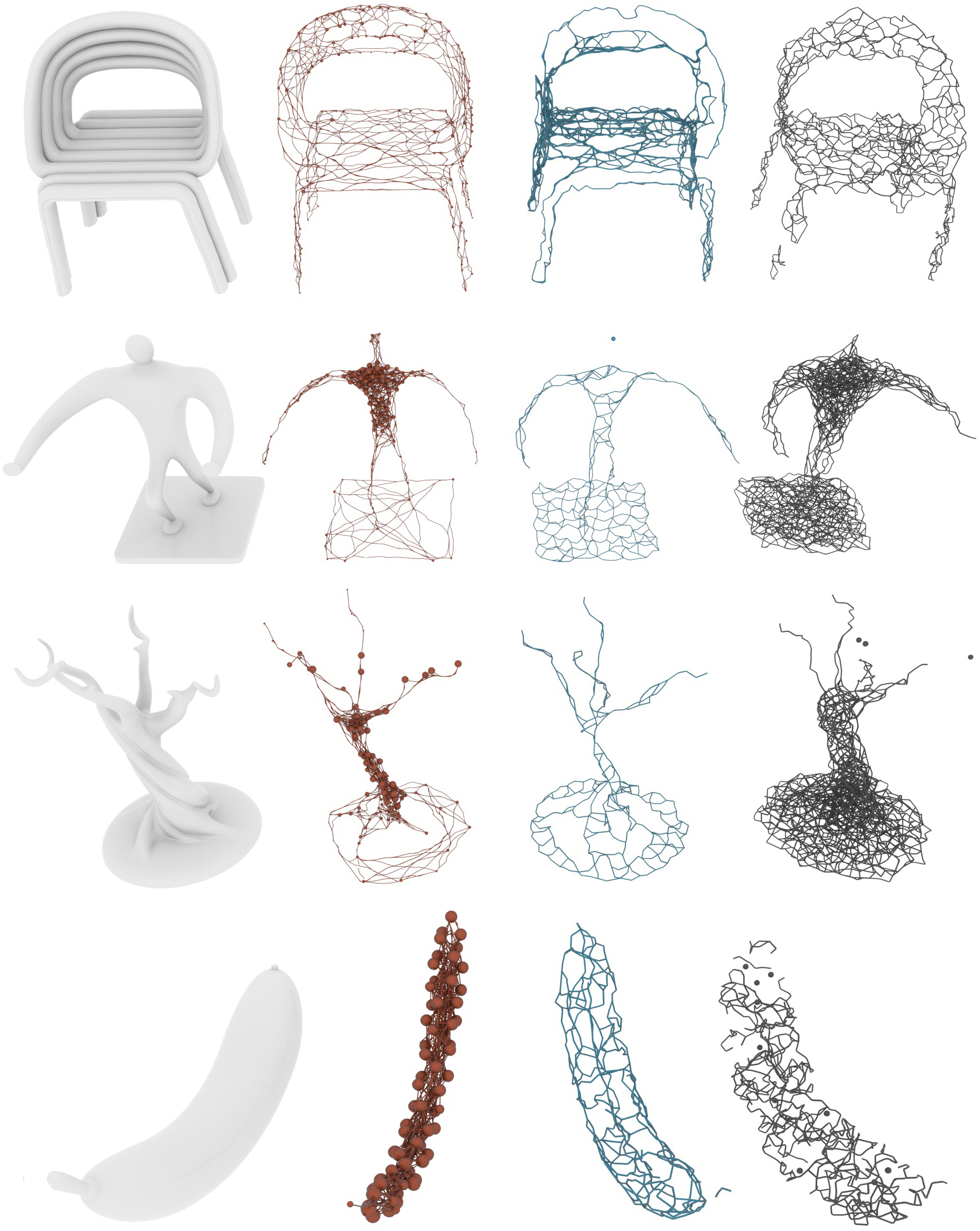}
    \caption{\textbf{Pure procedural algorithm comparison.} We show the results of the proposed 3D Topology Style Transfer approach. For each 3D input model (first column from left), we present its volumetric transformations based on the \textit{3D Network} topology (second column) and the \textit{Ghirigoro} topology (third column). We then compare the results of the Procedural Network algorithm (fourth column) that applies the volumetric transformation directly on the 3D input.
    }
    \label{fig:results}
\vspace{-10pt}
\end{figure}
\outNim{
\begin{figure*}[t]
    \centering
    \includegraphics[width=0.75\textwidth]{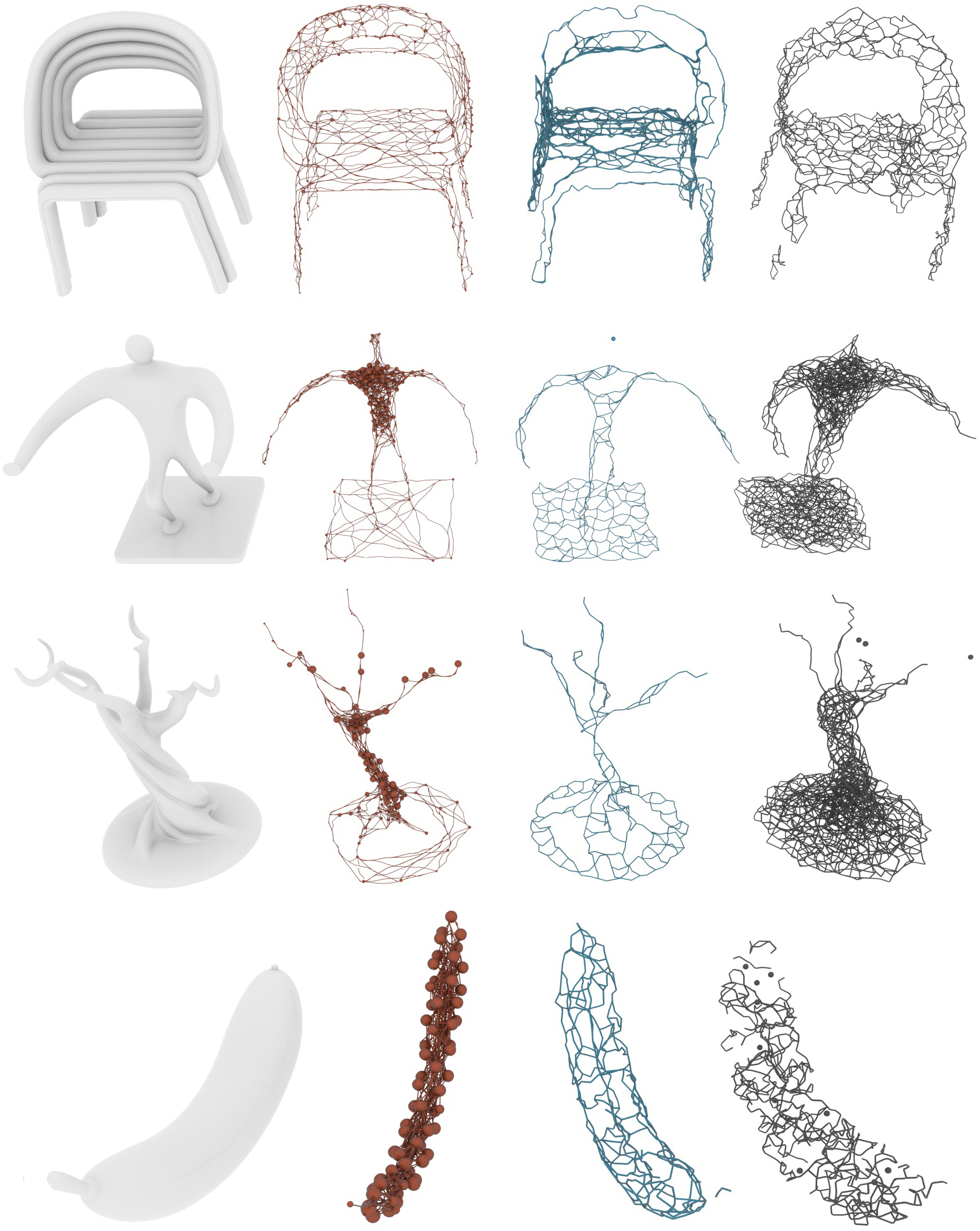}
    \caption{
    \small
    \textbf{Results.} We show the results of the proposed 3D Topology Style Transfer approach compared with the results of a stochastic algorithm (fourth in order) that directly applies the volumetric topology transformation on the 3D input. For each 3D input model (first model), we present its volumetric transformations based on the \textit{3D Network} topology (second in order) and the \textit{Ghirigoro} topology (third column). 
    }
    \label{fig:results}
\end{figure*}
}
As is evident in Figure~\ref{fig:results}, the baseline (fourth column, black model) looks quite messy. 
The baseline also requires a much higher node density to be able to capture the shape of the object. 
In comparison, using the Vox2Vox output a prior for the node and link distributions of the 3D Network (second column, red) results in much nicer results with much fewer nodes an links. 
Vox2Vox chooses node locations very economically and strategically, preserving the overall structure of the shape with much fewer nodes. 
Vox2Vox also does a very interesting job with the Ghirigoro style (third column, blue). 
The results all preserve the shape of the object very well, while using very few curved lines. 

Thus, our results show that it is possible to perform complex 3D topology style transfer using our Vox2Vox architecture, if we possess a good number of samples in the desired style, or are able to generate such samples.
The samples could have any arbitrary or random shape and the task of Vox2Vox is to figure out how to convert a given shape into that style.
Moreover, the resolution of the output can be modified, as long as the size of the features does not need to scale with the input size. 
Lastly, one could use other procedures to convert the Vox2Vox output to high quality 3D shapes, like our Procedural Network algorithm. 

\section{Conclusion and Future Direction}

In this paper, we presented a novel 3D-to-3D topology transfer paradigm based on transformations in 3D space. In particular, we built a 3D conditional GAN, Vox2Vox, that performs volumetric transformations to modify the internal structure of any 3D object, while maintaining its overall shape. 
We described our complete pipeline to apply our approach to two different topologies: the \textit{3D Network} and the \textit{Ghirigoro}. 
The results obtained by employing our methodology are novel and inspirational.
We compared the outputs of the pipeline while using or not the 3D-cGAN and found that using the Vox2Vox output as a prior distribution results in much nicer outcomes where features are placed in strategic positions in the 3D shape preserving its structural features. 
As a future direction, we plan to improve the 3D-to-3D topology transfer by given also the topology as a conditional input of the generative network. To do that, the machine learning algorithm has to learn itself the abstraction of the topology from a given 3D object.


\bibliographystyle{iccc}
\bibliography{mybib}

\begin{thebibliography}{}

\bibitem[\protect\citeauthoryear{Achlioptas \bgroup et al.\egroup
  }{2017}]{achlioptas2017learning}
Achlioptas, P.; Diamanti, O.; Mitliagkas, I.; and Guibas, L.
\newblock 2017.
\newblock Learning representations and generative models for 3d point clouds.
\newblock {\em arXiv preprint arXiv:1707.02392}.

\bibitem[\protect\citeauthoryear{Barab{\'a}si and
  Albert}{1999}]{barabasi1999emergence}
Barab{\'a}si, A.-L., and Albert, R.
\newblock 1999.
\newblock Emergence of scaling in random networks.
\newblock {\em science} 286(5439):509--512.

\bibitem[\protect\citeauthoryear{Briot, Hadjeres, and
  Pachet}{2017}]{briot2017deep}
Briot, J.-P.; Hadjeres, G.; and Pachet, F.
\newblock 2017.
\newblock Deep learning techniques for music generation-a survey.
\newblock {\em arXiv preprint arXiv:1709.01620}.

\bibitem[\protect\citeauthoryear{Brock \bgroup et al.\egroup
  }{2016}]{brock2016generative}
Brock, A.; Lim, T.; Ritchie, J.~M.; and Weston, N.
\newblock 2016.
\newblock Generative and discriminative voxel modeling with convolutional
  neural networks.
\newblock {\em arXiv preprint arXiv:1608.04236}.

\bibitem[\protect\citeauthoryear{Brock, Donahue, and
  Simonyan}{2018}]{brock2018large}
Brock, A.; Donahue, J.; and Simonyan, K.
\newblock 2018.
\newblock Large scale gan training for high fidelity natural image synthesis.
\newblock {\em arXiv preprint arXiv:1809.11096}.

\bibitem[\protect\citeauthoryear{Dehmamy, Milanlouei, and
  Barab\'asi}{2018}]{dehmamy2018structural}
Dehmamy, N.; Milanlouei, S.; and Barab\'asi, A.-L.
\newblock 2018.
\newblock Structural transition in physical networks.
\newblock {\em {Nature} (to appear)}.

\bibitem[\protect\citeauthoryear{Dumoulin, Shlens, and
  Kudlur}{2017}]{dumoulin2017learned}
Dumoulin, V.; Shlens, J.; and Kudlur, M.
\newblock 2017.
\newblock A learned representation for artistic style.
\newblock {\em Proc. of ICLR}.

\bibitem[\protect\citeauthoryear{Erd{\H{o}}s and
  R{\'e}nyi}{1960}]{erdHos1960evolution}
Erd{\H{o}}s, P., and R{\'e}nyi, A.
\newblock 1960.
\newblock On the evolution of random graphs.
\newblock {\em Publ. Math. Inst. Hung. Acad. Sci} 5(1):17--60.

\bibitem[\protect\citeauthoryear{Gatys, Ecker, and
  Bethge}{2016}]{gatys2016image}
Gatys, L.~A.; Ecker, A.~S.; and Bethge, M.
\newblock 2016.
\newblock Image style transfer using convolutional neural networks.
\newblock In {\em Proceedings of the IEEE Conference on Computer Vision and
  Pattern Recognition},  2414--2423.

\bibitem[\protect\citeauthoryear{Goodfellow \bgroup et al.\egroup
  }{2014}]{goodfellow2014generative}
Goodfellow, I.; Pouget-Abadie, J.; Mirza, M.; Xu, B.; Warde-Farley, D.; Ozair,
  S.; Courville, A.; and Bengio, Y.
\newblock 2014.
\newblock Generative adversarial nets.
\newblock In {\em Advances in neural information processing systems},
  2672--2680.

\bibitem[\protect\citeauthoryear{Ha and Eck}{2017}]{ha2017neural}
Ha, D., and Eck, D.
\newblock 2017.
\newblock A neural representation of sketch drawings.
\newblock {\em arXiv preprint arXiv:1704.03477}.

\bibitem[\protect\citeauthoryear{Hart, Nilsson, and
  Raphael}{1968}]{hart1968formal}
Hart, P.~E.; Nilsson, N.~J.; and Raphael, B.
\newblock 1968.
\newblock A formal basis for the heuristic determination of minimum cost paths.
\newblock {\em IEEE transactions on Systems Science and Cybernetics}
  4(2):100--107.

\bibitem[\protect\citeauthoryear{Huang \bgroup et al.\egroup
  }{2017}]{huang2017real}
Huang, H.; Wang, H.; Luo, W.; Ma, L.; Jiang, W.; Zhu, X.; Li, Z.; and Liu, W.
\newblock 2017.
\newblock Real-time neural style transfer for videos.
\newblock In {\em 2017 IEEE Conference on Computer Vision and Pattern
  Recognition (CVPR)},  7044--7052.
\newblock IEEE.

\bibitem[\protect\citeauthoryear{Huang, Kalogerakis, and
  Marlin}{2015}]{huang2015analysis}
Huang, H.; Kalogerakis, E.; and Marlin, B.
\newblock 2015.
\newblock Analysis and synthesis of 3d shape families via deep-learned
  generative models of surfaces.
\newblock In {\em Computer Graphics Forum}, volume~34,  25--38.
\newblock Wiley Online Library.

\bibitem[\protect\citeauthoryear{Isola \bgroup et al.\egroup
  }{2017}]{isola2017image}
Isola, P.; Zhu, J.-Y.; Zhou, T.; and Efros, A.~A.
\newblock 2017.
\newblock Image-to-image translation with conditional adversarial networks.
\newblock {\em arXiv preprint}.

\bibitem[\protect\citeauthoryear{Johnson, Alahi, and
  Fei-Fei}{2016}]{johnson2016perceptual}
Johnson, J.; Alahi, A.; and Fei-Fei, L.
\newblock 2016.
\newblock Perceptual losses for real-time style transfer and super-resolution.
\newblock In {\em European Conference on Computer Vision},  694--711.
\newblock Springer.

\bibitem[\protect\citeauthoryear{Kalogerakis \bgroup et al.\egroup
  }{2012}]{kalogerakis2012probabilistic}
Kalogerakis, E.; Chaudhuri, S.; Koller, D.; and Koltun, V.
\newblock 2012.
\newblock A probabilistic model for component-based shape synthesis.
\newblock {\em ACM Transactions on Graphics (TOG)} 31(4):55.

\bibitem[\protect\citeauthoryear{Kato, Ushiku, and
  Harada}{2018}]{kato2018neural}
Kato, H.; Ushiku, Y.; and Harada, T.
\newblock 2018.
\newblock Neural 3d mesh renderer.
\newblock In {\em Proceedings of the IEEE Conference on Computer Vision and
  Pattern Recognition},  3907--3916.

\bibitem[\protect\citeauthoryear{Li \bgroup et al.\egroup }{2018}]{li2018point}
Li, C.-L.; Zaheer, M.; Zhang, Y.; Poczos, B.; and Salakhutdinov, R.
\newblock 2018.
\newblock Point cloud gan.
\newblock {\em arXiv preprint arXiv:1810.05795}.

\bibitem[\protect\citeauthoryear{Li \bgroup et al.\egroup
  }{2019}]{li2019synthesizing}
Li, X.; Dong, Y.; Peers, P.; and Tong, X.
\newblock 2019.
\newblock Synthesizing 3d shapes from silhouette image collections using
  multi-projection generative adversarial networks.
\newblock In {\em Proceedings of the IEEE Conference on Computer Vision and
  Pattern Recognition},  5535--5544.

\bibitem[\protect\citeauthoryear{Liu \bgroup et al.\egroup
  }{2015}]{liu2015style}
Liu, T.; Hertzmann, A.; Li, W.; and Funkhouser, T.
\newblock 2015.
\newblock Style compatibility for 3d furniture models.
\newblock {\em ACM Transactions on Graphics (TOG)} 34(4):85.

\bibitem[\protect\citeauthoryear{Lun, Kalogerakis, and
  Sheffer}{2015}]{lun2015elements}
Lun, Z.; Kalogerakis, E.; and Sheffer, A.
\newblock 2015.
\newblock Elements of style: learning perceptual shape style similarity.
\newblock {\em ACM Transactions on Graphics (TOG)} 34(4):84.

\bibitem[\protect\citeauthoryear{Ma \bgroup et al.\egroup
  }{2014}]{ma2014analogy}
Ma, C.; Huang, H.; Sheffer, A.; Kalogerakis, E.; and Wang, R.
\newblock 2014.
\newblock Analogy-driven 3d style transfer.
\newblock In {\em Computer Graphics Forum}, volume~33,  175--184.
\newblock Wiley Online Library.

\bibitem[\protect\citeauthoryear{Mazeika and
  Whitehead}{2018}]{mazeika2018towards}
Mazeika, J., and Whitehead, J.
\newblock 2018.
\newblock Towards 3d neural style transfer.
\newblock In {\em AIIDE Workshops}.

\bibitem[\protect\citeauthoryear{Mirza and
  Osindero}{2014}]{mirza2014conditional}
Mirza, M., and Osindero, S.
\newblock 2014.
\newblock Conditional generative adversarial nets.
\newblock {\em arXiv preprint arXiv:1411.1784}.

\bibitem[\protect\citeauthoryear{Nguyen \bgroup et al.\egroup
  }{2012}]{nguyen20123d}
Nguyen, C.~H.; Ritschel, T.; Myszkowski, K.; Eisemann, E.; and Seidel, H.-P.
\newblock 2012.
\newblock 3d material style transfer.
\newblock In {\em Computer Graphics Forum}, volume~31,  431--438.
\newblock Wiley Online Library.

\bibitem[\protect\citeauthoryear{Ongun and Temizel}{2018}]{ongun2018paired}
Ongun, C., and Temizel, A.
\newblock 2018.
\newblock Paired 3d model generation with conditional generative adversarial
  networks.
\newblock {\em arXiv preprint arXiv:1808.03082}.

\bibitem[\protect\citeauthoryear{Qi \bgroup et al.\egroup
  }{2016}]{qi2016volumetric}
Qi, C.~R.; Su, H.; Nie{\ss}ner, M.; Dai, A.; Yan, M.; and Guibas, L.~J.
\newblock 2016.
\newblock Volumetric and multi-view cnns for object classification on 3d data.
\newblock In {\em Proceedings of the IEEE conference on computer vision and
  pattern recognition},  5648--5656.

\bibitem[\protect\citeauthoryear{Radford, Metz, and
  Chintala}{2015}]{radford2015unsupervised}
Radford, A.; Metz, L.; and Chintala, S.
\newblock 2015.
\newblock Unsupervised representation learning with deep convolutional
  generative adversarial networks.
\newblock {\em arXiv preprint arXiv:1511.06434}.

\bibitem[\protect\citeauthoryear{Ronneberger, Fischer, and
  Brox}{2015}]{ronneberger2015u}
Ronneberger, O.; Fischer, P.; and Brox, T.
\newblock 2015.
\newblock U-net: Convolutional networks for biomedical image segmentation.
\newblock In {\em International Conference on Medical image computing and
  computer-assisted intervention},  234--241.
\newblock Springer.

\bibitem[\protect\citeauthoryear{Shin, Fowlkes, and
  Hoiem}{2018}]{shin2018pixels}
Shin, D.; Fowlkes, C.~C.; and Hoiem, D.
\newblock 2018.
\newblock Pixels, voxels, and views: A study of shape representations for
  single view 3d object shape prediction.
\newblock In {\em Proceedings of the IEEE conference on computer vision and
  pattern recognition},  3061--3069.

\bibitem[\protect\citeauthoryear{Soltani \bgroup et al.\egroup
  }{2017}]{soltani2017synthesizing}
Soltani, A.~A.; Huang, H.; Wu, J.; Kulkarni, T.~D.; and Tenenbaum, J.~B.
\newblock 2017.
\newblock Synthesizing 3d shapes via modeling multi-view depth maps and
  silhouettes with deep generative networks.
\newblock In {\em The IEEE conference on computer vision and pattern
  recognition (CVPR)}, volume~3, ~4.

\bibitem[\protect\citeauthoryear{Su \bgroup et al.\egroup }{2015}]{su2015multi}
Su, H.; Maji, S.; Kalogerakis, E.; and Learned-Miller, E.
\newblock 2015.
\newblock Multi-view convolutional neural networks for 3d shape recognition.
\newblock In {\em Proceedings of the IEEE international conference on computer
  vision},  945--953.

\bibitem[\protect\citeauthoryear{Sumner and
  Popovi{\'c}}{2004}]{sumner2004deformation}
Sumner, R.~W., and Popovi{\'c}, J.
\newblock 2004.
\newblock Deformation transfer for triangle meshes.
\newblock In {\em ACM Transactions on Graphics (TOG)}, volume~23,  399--405.
\newblock ACM.

\bibitem[\protect\citeauthoryear{Wu \bgroup et al.\egroup
  }{2016}]{wu2016learning}
Wu, J.; Zhang, C.; Xue, T.; Freeman, B.; and Tenenbaum, J.
\newblock 2016.
\newblock Learning a probabilistic latent space of object shapes via 3d
  generative-adversarial modeling.
\newblock In {\em Advances in Neural Information Processing Systems},  82--90.

\bibitem[\protect\citeauthoryear{Xie \bgroup et al.\egroup
  }{2018}]{xie2018learning}
Xie, J.; Zheng, Z.; Gao, R.; Wang, W.; Zhu, S.-C.; and Wu, Y.~N.
\newblock 2018.
\newblock Learning descriptor networks for 3d shape synthesis and analysis.
\newblock In {\em Proceedings of the IEEE Conference on Computer Vision and
  Pattern Recognition},  8629--8638.

\bibitem[\protect\citeauthoryear{Xu \bgroup et al.\egroup }{2010}]{xu2010style}
Xu, K.; Li, H.; Zhang, H.; Cohen-Or, D.; Xiong, Y.; and Cheng, Z.-Q.
\newblock 2010.
\newblock Style-content separation by anisotropic part scales.
\newblock In {\em ACM Transactions on Graphics (TOG)}, volume~29,  184.
\newblock ACM.

\bibitem[\protect\citeauthoryear{Zhu \bgroup et al.\egroup
  }{2017}]{CycleGAN2017}
Zhu, J.-Y.; Park, T.; Isola, P.; and Efros, A.~A.
\newblock 2017.
\newblock Unpaired image-to-image translation using cycle-consistent
  adversarial networkss.
\newblock In {\em Computer Vision (ICCV), 2017 IEEE International Conference
  on}.

\end{thebibliography}

\end{document}